\documentclass[12pt]{iopart}

\usepackage{booktabs}
\usepackage{graphics}
\usepackage{graphicx}
\usepackage{acronym}
\usepackage{amsmath}
\usepackage{subcaption}
\usepackage{xcolor}

\acrodef{aoa}[AoA]{Angle of Arrival}
\acrodef{aod}[AoD]{Angle of Departure}
\acrodef{cir}[CIR]{Channel Impulse Response}
\acrodef{cnn}[CNN]{Convolutional Neural Network}
\acrodef{eflop}[EFLOP]{Effective Floating Point Operation}

\acrodef{dft}[DFT]{Discrete Fourier Transform}
\acrodef{dl}[DL]{Deep Learning}
\acrodef{elu}[ELU]{Exponential-Linear Unit}
\acrodef{fmcw}[FMCW]{Frequency-Modulated Continuous-Wave}
\acrodef{fov}[FOV]{Field-of-View}
\acrodef{ft}[FT]{Fourier Transform}
\acrodef{har}[HAR]{Human Activity Recognition}
\acrodef{if}[IF]{Intermediate Frequency}
\acrodef{los}[LOS]{Line-of-Sight}
\acrodef{mae}[MAE]{Mean Absolute Error}
\acrodef{mse}[MSE]{Mean Squared Error}
\acrodef{md}[$\mu$D]{micro-Doppler}
\acrodef{rd}[RD]{range-Doppler}
\acrodef{mmwave}[mmWave]{Millimeter-Wave}
\acrodef{nn}[NN]{Neural Network}
\acrodef{ofdm}[OFDM]{Orthogonal Frequency Division Multiplexing}
\acrodef{rf}[RF]{Radio Frequency}
\acrodef{relu}[ReLU]{Rectified Linear Unit}
\acrodef{ann}[ANN]{Artificial Neural Network}
\acrodef{rnn}[RNN]{Recurrent Neural Network}
\acrodef{rmse}[RMSE]{Root MSE}
\acrodef{siso}[SISO]{Single Input Single Output}
\acrodef{snr}[SNR]{Signal-to-Noise Ratio}
\acrodef{stft}[STFT]{Short Time Fourier Transform}
\acrodef{snn}[SNN]{Spiking Neural Network}
\acrodef{tf}[TF]{Time-Frequency}
\acrodef{gru}[GRU]{Gated Recurrent Unit}
\acrodef{flops}[FLOPs]{Floating Point Operations}
\acrodef{dvs}[DVS]{Dynamic Vision Sensor}
\acrodef{lstm}[LSTM]{Long Short Term Memory}

\acrodef{elu}[ELU]{Exponential Linear Unit}
\acrodef{mlp}[MLP]{Multi Layer Perceptron}

\acrodef{lif}[LIF]{Leaky Integrate and Fire}

\begin{document}

\title[Spatiotemporal Radar Gesture Recognition with Hybrid Spiking Neural Networks]{Spatiotemporal Radar Gesture Recognition with Hybrid Spiking Neural Networks: Balancing Accuracy and Efficiency}

\author{Riccardo Mazzieri$^1$, Eleonora Cicciarella$^1$, Jacopo Pegoraro$^1$, Federico Corradi$^2$, and Michele Rossi$^{1,3}$}

\address{$^1$ Department of Information Engineering, University of Padova, Padova 35131, Italy}
\address{$^2$ Department of Electrical Engineering, Eindhoven University of Technology, Eindhoven, The Netherlands}
\address{$^3$ Department of Mathematics ``Tullio Levi-Civita", University of Padova, Padova 35121, Italy}
\ead{riccardo.mazzieri@phd.unipd.it}

\vspace{10pt}
\begin{indented}
\item[]June 2025
\end{indented}


\begin{abstract}
Radar-based Human Activity Recognition (HAR) offers privacy and robustness over camera-based methods, yet remains computationally demanding for edge deployment.
We present the first use of Spiking Neural Networks (SNNs) for radar-based HAR on aircraft marshalling signal classification.
Our novel hybrid architecture combines convolutional modules for spatial feature extraction with Leaky Integrate-and-Fire (LIF) neurons for temporal processing, inherently capturing gesture dynamics. The model reduces trainable parameters by 88\% with under 1\% accuracy loss compared to baselines, and generalizes well to the Soli gesture dataset. Through systematic comparisons with Artificial Neural Networks, we demonstrate the trade-offs of spiking computation in terms of accuracy, latency, memory, and energy, establishing SNNs as an efficient and competitive solution for radar-based HAR.

\end{abstract}

%
%
%
%
%

\section{Introduction}
Recent progress in \ac{dl} techniques, along with advances in radar technologies, has caused a growing interest in the research field of radar-based \ac{har}. In parallel, technological improvements, such as the use of higher carrier frequencies, the integration of antennas within chip packages, and higher-resolution ADCs, are making radar sensors increasingly compact and cost-efficient, thus enabling their pervasive deployment at the edge.

By processing the backscattered radio signal from human targets through dedicated signal processing pipelines, it is possible to extract motion-related quantities such as range, radial velocity, and angle of arrival of the targets. This, complemented with the advanced data-driven feature extraction capabilities of modern neural networks, enables a wide range of applications such as hand gesture recognition \cite{tsang2021radar}, vital signs monitoring \cite{8695699}, fall detection \cite{7907206}, or automatic systems for tracking and identification \cite{pegoraro2020multiperson}.
Unlike more conventional camera-based solutions, radar sensing preserves user privacy and is inherently robust to adverse environmental conditions, such as poor lighting or the presence of fog or smoke. For these reasons, radar technologies have the potential to serve as an alternative or complementary modality for \ac{har}, compared to more traditional camera-based vision systems.

Additionally, edge computing aims to shift a significant portion of the computational workload and data storage to the edge of the network, thereby reducing the energy demands typically associated with cloud-based solutions. This paradigm shift requires the development of efficient neural architectures that can operate within the memory and power constraints of edge devices and embedded systems.
As a result, research of lightweight architectures capable of retaining competitive performance while significantly reducing resource consumption is becoming more and more relevant in this context. For this reason, alternative paradigms such as \acp{snn}, that offer sparse and event-driven computations with higher energy efficiency, are gaining increasing attention in this domain~\cite{zhang2024spiking,corradi2025-opportunities}.

Most existing works on radar-based \ac{har} employ representations such as \ac{md} spectrograms \cite{skaria2019hand} or range-doppler maps \cite{tsang2021radar} as input to the neural network. This choice of image-like input is motivated by the effectiveness of neural architectures such as \acp{cnn} in extracting fine grained visual features and still retaining good generalization capabilities. However, this type of single-frame input is less suited for networks employing recurrent architectures, such as \ac{lstm} \cite{lstm_graves2012long} or \ac{gru} \cite{gru_cho2014properties} cells, or \ac{snn}s, which on the other hand are inherently designed for temporal correlation modeling, and thus require a sequential, time-varying input representation. 

\begin{figure*}
    \includegraphics[width=\linewidth]{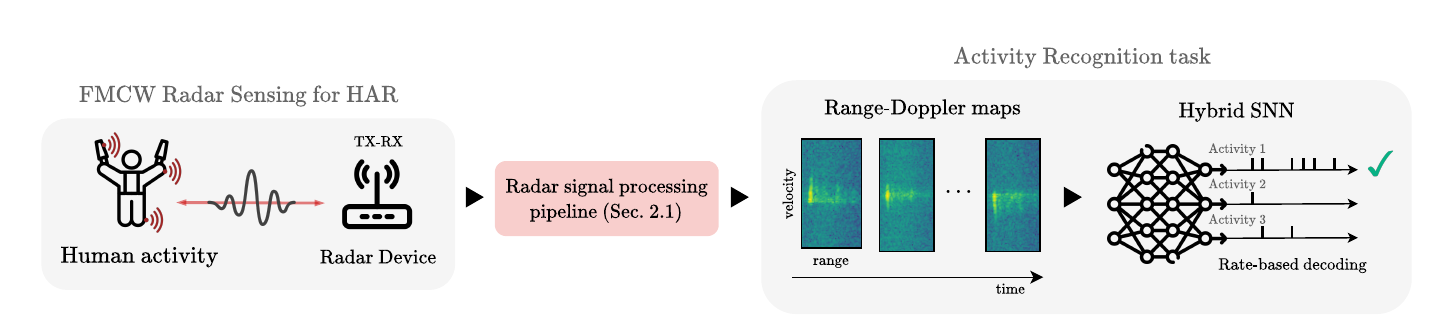}
    \caption{System pipeline for radar-based \ac{har}. First, a \ac{fmcw} radar device performs contactless sensing of the human gesture. By leveraging key information in the back-scattered radio wave, a dedicated signal processing pipeline (Section \ref{sec:radar_sigproc}) is employed to extract key quantities such as the distance and velocity of the subjects' body parts in the form of Range-Doppler map sequences. This sequential representation is processed by a hybrid \ac{snn} to perform training and inference for the activity recognition task. }
    \label{fig:radar_snn_pipeline}
\end{figure*}

In this work, we investigate the use of \acp{snn} for radar-based \ac{har}, building a complete system pipeline going from pre-processing, network training, validation and model complexity optimization. After building a sequential \ac{fmcw} radar input representation, we design and evaluate several neural architectures, including a hybrid architecture based both on \ac{ann} and \ac{snn} modules. 
All proposed architectures are trained and validated in the task of classification of gestures in a public dataset of aircraft signal processing~\cite{10149465}. This task is particularly challenging, as marshalling signals unfold over a few seconds and requires the model to capture temporal correlations throughout the sequence. Figure \ref{fig:radar_snn_pipeline} shows a complete outline of our system pipeline for radar-based \ac{har} with \acp{snn}.

The contributions of this work are the following.
\begin{enumerate}
    \item Unlike existing solutions in the literature, we approach the problem by using a time-stepped representation of the input, and we present a set of network architectures that explicitly model the time correlation of the gestures' spatial features. This approach enables us to surpass the state-of-the-art performance (accuracy) on the aricraft marshalling dataset. 
    \item This is the first work to apply \acp{snn} to this dataset. We introduce a hybrid architecture that combines convolutional and spiking layers for aircraft marshalling signal classification. By leveraging time-stepped input and LIF neurons, our model inherently captures temporal correlations. We further demonstrate its flexibility through experimental validation on the Soli dataset.
    \item We systematically assess \ac{snn}-based architectures for radar-based \ac{har}, analyzing accuracy, memory, latency and energy efficiency.
    Our hybrid model exposes clear trade-offs between performance and complexity, making it well-suited for edge deployment. Remarkably, it reduces trainable parameters by $88\%$  
    compared to baseline approaches, with 
    drop in accuracy of less than $1\%$.
\end{enumerate}


\section{Previous Work}

\subsection{Radar Signal Processing}
\label{sec:radar_sigproc}

\ac{fmcw} radars operate by transmitting \textit{chirp} waveforms that linearly sweep a bandwidth $B$ from a minimum frequency $f_{0}$ to a maximum frequency $f_{1}$, with $B = f_{0} - f_{1}$.
As a result, the phase of the transmitted signal can be written as $\phi_{\rm tx}(t) = 2\pi (f_{0} + \kappa t) t$ where $\kappa = B/T_{\rm c}$ 
and
$T_{\rm c}$ is the duration of the chirp.

To increase the \ac{snr} and enable Doppler estimation, \ac{fmcw} radars typically transmit bursts of $M$ chirp signals, indexed by $m=0, \dots, M-1$, denoted by \textit{radar frames}, every $T_{\rm f}$ seconds (\textit{slow time} dimension).
The transmitted radar frames are scattered by objects and obstacles in the environment, so that the received signals contain information about their location and movement speed.
To extract such information, the radar receiver performs mixing of the received signal with the transmitted one to remove the carrier frequency.
Then, it samples each chirp in the burst with period $T_{\rm s}$, where samples are indexed by $n=0,\dots, N-1$, called \textit{fast time} dimension.
The resulting phase of the radar received signal, considering a single target located at $R$ meters from the radar, and moving with radial velocity $v$, is~\cite{pegoraro2020multiperson}
\begin{equation}\label{eq:rx-phase}
\phi_{\rm rx}(n, m) \approx 2\pi \left(\frac{2 R f_{0}}{c} +\frac{2v f_{0} }{c} m T_{\rm f}  + \frac{2 \kappa R}{c} n T_{\rm s} \right),
\end{equation}
where the approximation is customary and neglects second-order terms.
From Eq.~(\ref{eq:rx-phase}), a \ac{dft} along the fast time dimension yields a peak at $2\kappa R T_{\rm s} / c$, while a \ac{dft} along the slow time dimension yields a peak at $2vf_0T_{\rm f} /c$.
From the peak locations, $R$ and $v$ can be estimated knowing the radar transmission parameters.
Due to the application of the \ac{dft}, the resolution on the estimate of the distance, $\Delta R$, and velocity of the target are finite and equal to~\cite{pegoraro2020multiperson}
\begin{equation}
\Delta R =\frac{c}{2B}, \quad \quad \Delta v = \frac{c}{2f_0 MT_{\rm f}}.
\end{equation}
In the case of multiple targets, typically \ac{fmcw} radar signal processing collects the received signal samples in a matrix of shape $N\times M$. A 2D \ac{dft} is then applied along fast and slow time dimensions, and the magnitude square of the result is taken to identify the peaks corresponding to the multiple targets. 
The resulting matrix is called \ac{rd} map, where the location of each peak corresponds to the distance and velocity parameters of a given target.

\subsection{Radar-based \ac{har} with \ac{snn}s}

The literature on radar-based \ac{har} is broad and has traditionally relied on deep neural networks to automatically extract spatio-temporal features of human movements from radar representations, such as \ac{rd} maps and \ac{md} spectrograms \cite{chen2000analysis}. While these methods have demonstrated excellent performance, the reliance on large-scale convolutional and recurrent architectures results in high computational complexity and energy consumption, preventing their deployment on embedded or edge devices. Given the increasing interest in radar technologies, optimizing recognition models for energy efficiency has become a priority. In this context, \ac{snn}s have attracted significant attention due to their event-driven and sparse processing paradigm, which carries the potential for orders-of-magnitude energy savings when deployed on dedicated neuromorphic hardware \cite{corradi2025-opportunities}.

In one of the first relevant works to explore radar-based gesture recognition with \ac{snn} \cite{tsang2021radar}, the authors process FMCW millimeter-wave radar signals (\ac{rd} and \ac{md} maps) by encoding them into spike trains and feeding them into a Liquid State Machine (LSM). Despite using fewer than 1000 neurons, their system achieves high accuracy on two public hand-gesture datasets. 
Another early contribution \cite{arsalan2021radar} introduces a \ac{snn} trained with the surrogate-gradient method~\cite{eshraghian2023training} for radar-based gesture recognition. 
In \cite{arsalan2021radar}, the radar signals are converted into \ac{rd} images and the temporal sequences of range–Doppler vectors are extracted across frames, embedding temporal dynamics into the representation. This encoding is particularly suited to spiking architectures, as it naturally preserves time information. The proposed network is a lightweight architecture composed of dense layers interleaved with \ac{lif} neurons, trained with a differentiable SoftLIF activation and tested with true spiking LIF units. 
In \cite{arsalan2022spiking}, the same authors propose a spiking network architecture for gesture recognition working directly from raw ADC radar signals, avoiding altogether the need for conventional radar signal-processing steps such as 2D FFTs. Their results, obtained on the same hand-gesture dataset, show that \acp{snn} can emulate the functionality of traditional radar signal processing chains within the network itself, paving the way for highly efficient end-to-end neuromorphic implementations.

Other works have explored hybrid solutions that combine spiking and conventional neural processing. For example, \cite{gerhards2023hybrid} introduces a hybrid SNN architecture trained from scratch with a surrogate-gradient training approach on a four-class radar gesture dataset. Their architecture is composed of both a CNN for spatial features modeling, and SNN modules for temporal correlation modeling, showing the feasibility of end-to-end learning with small datasets.
Similarly, in \cite{wu2025efficient}, the authors propose an efficient radar-based gesture recognition framework that integrates an enhanced Gaussian Mixture Model (GMM) for denoising and pre-processing, with a hybrid \ac{snn} composed of a convolutional SNN (CSNN) for spatial–temporal feature extraction and LSTM units for long-term temporal modeling. 

In the specific case of the Aircraft Marshalling Signals dataset, which is the focus of this work, only a limited number of studies have been reported in the literature. In \cite{10149465}, the dataset was first introduced together with a spike-encoding scheme for radar representations. The authors propose simple baseline models for both radar-only processing and sensor fusion with \ac{dvs} data. Their radar baseline employs a single \ac{rd} map per gesture, condensed into a single frame and processed using a pretrained CNN backbone. 
Later, \cite{10734929} investigated more advanced ANN architectures and proposed an improved radar pre-processing pipeline, improving the state-of-the-art accuracy. Nevertheless, their approach also relies on single-frame \ac{rd} maps, preventing explicit modeling of the temporal correlation present in the input gestures, which is a crucial aspect for spiking-based approaches that naturally encode and exploit temporal information.

The above works prove the viability of radar-based \ac{har} with \acp{snn}. Early contributions demonstrate the feasibility of compact spiking architectures, while more recent works explore hybrid designs. However, in the case of the Aircraft Marshalling Signals dataset, existing studies do not investigate the use of \ac{snn}, and rely on a simplified data representation and pre-processing pipeline, leaving room for additional improvements and experimentation. Moreover, the current literature does not adequately examine the unique advantages, limitations, and trade-offs of employing spiking architectures for radar-based \ac{har}, especially when compared to more conventional approaches in the context of resource-constrained edge applications.

\section{Proposed Method}


\subsection{Dataset overview}
\label{subsec:dataset}

All models presented in this work are trained and evaluated on the Aircraft Marshaling Signals Dataset~\cite{10149465}, a publicly available multimodal dataset combining low-resolution radar and event-based vision data for whole-body gesture recognition. The dataset contains \mbox{$M=11$} gesture classes, comprising ten aircraft marshaling signals: \textit{turn right, turn left, straight ahead, stop engines, start engines, slow down, move back v1, move back v2, move ahead, emergency stop} and one background \textit{none} class, where no subject was present in front of the sensors.

Data were collected using an 8~GHz FMCW SISO radar (750~MHz bandwidth)~\cite{liu2019ultralow} and a \ac{dvs} camera~\cite{brandli2014240}, precisely synchronized at the chirp level to enable cross-modal alignment. The recordings were made in three indoor environments at eight discrete distances from the sensing platform. 

\begin{figure*}
    \centering
    \includegraphics[width=0.83\linewidth]{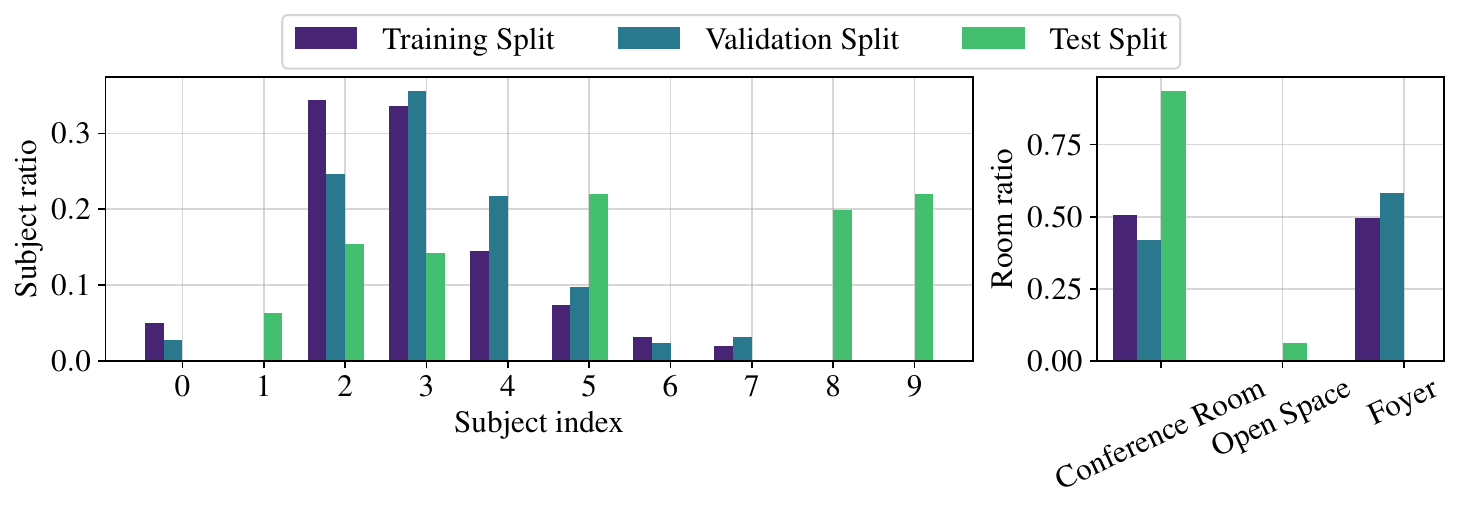}
    \caption{Subject and room distribution in the Aircraft Marshalling Signals dataset. The test split differs considerably from the training split, featuring three different subjects and a different environment.}
    \label{fig:dataset_statistics}
\end{figure*}

A notable challenge of this dataset is the deliberate \textit{domain shift} between training and testing data, featuring new, unseen subjects and environmental conditions in the test set, as shown in Figure~\ref{fig:dataset_statistics}.
This setup requires models to generalize across subject-specific motion patterns, varying background clutter, and changes in reflection geometry. 
Consequently, our results show that, while validation accuracies during training can reach high values, test-set metrics are consistently lower due to the increased difficulty of cross-subject and environment generalization. 

This combination of multimodal sensing, realistic domain shift, and motion complexity makes the Aircraft Marshaling Signals Dataset a challenging and relevant benchmark for evaluating gesture recognition models in real-world conditions.

\subsection{Data Pre-processing}

The Aricraft Marshalling Signals dataset is provided in the form of raw ADC data coming from an \ac{fmcw} radar sensor from Imec~\cite{liu2019ultralow}. As a consequence, pre-processing steps are required, each entailing differences in the quality of the resulting \ac{rd} maps.
For example, in previous works \cite{10149465, 10734929}, the network is provided with a single frame representation of the gesture, consisting of a \ac{rd} map extracted from a single radar frame. 
We argue that this is a simplistic approach since it compresses all the information of the gesture into a single image.
This approach does not model the temporal correlation of the gesture movements explicitly. Additionally, this representation is not suitable for \acp{snn}, which are designed for modeling time-varying input signals. 

In contrast with the existing approaches, in this work we adopt a sequential representation of the gesture, consisting of a sequence of \ac{rd} maps, obtained from different subsequent overlapping radar frames. This approach enables splitting the processing of the input sequence into a first spatial feature extraction phase and then a second temporal correlation processing phase, a technique that has proven effective in the deep learning literature \cite{gerhards2023hybrid, pegoraro2021real}.

Given the raw ADC data, different hyperparameter configurations are possible to obtain sequences of \ac{rd} maps. 
For example, by varying the radar frame length, we are determining the velocity resolution of the radar. 
With shorter radar frames, we are limiting the velocity resolution of the sensor, thus reducing the amount of detectable doppler frequencies. 
By doing the opposite, the obtained sequence will result in more overlapping frames, which will in turn result in more correlation between subsequent frames, together with blurry patterns in the resulting \ac{rd} map. 
To determine suitable values for input sequence length, radar frame length, and number of chirps, we systematically evaluated multiple hyperparameter settings.
In this work, each sample consists of sequences of $L=15$ radar frames,
each containing $256$ chirps with an overlap between subsequent frames of $146$ chirps, resulting in a total sequence duration of $3$~seconds. 

\begin{figure*}
    \centering
    \includegraphics[width=\linewidth]{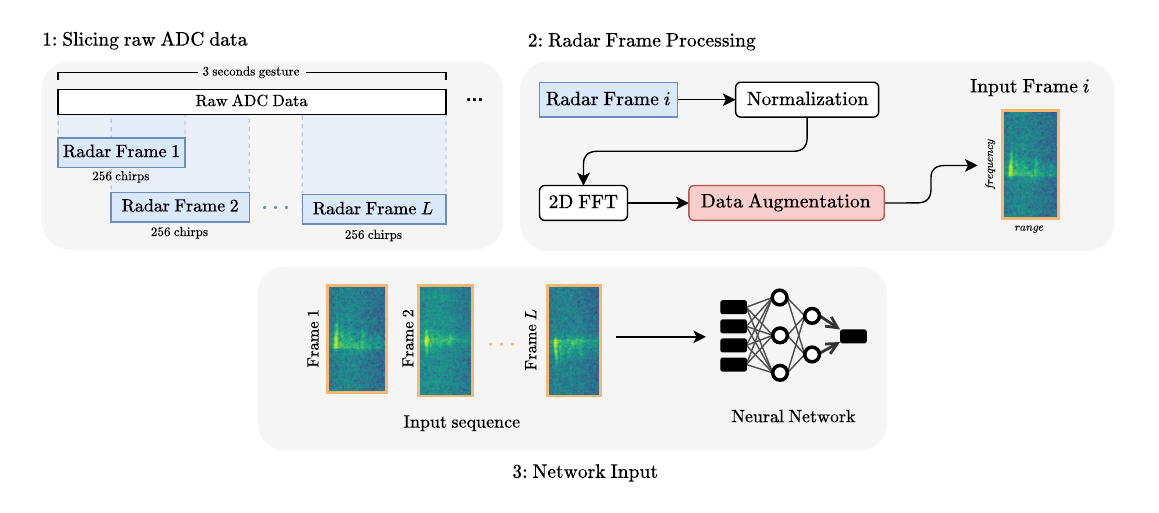}
    \caption{Outline of the data pre-processing pipeline. The raw ADC samples of the radar signal after mixing are first sliced into overlapping radar frames, each consisting of 256 chirps. Each frame then undergoes normalization and is transformed both along the fast and slow time dimensions by a 2D FFT. After that, a data augmentation procedure is applied to virtually increase the size of the dataset to prevent overfitting.}
    \label{fig:preproc}
\end{figure*}

Raw ADC data are first normalized to filter out static targets, after which a 2D FFT is applied to obtain the \ac{rd} maps. 
To improve generalization and mitigate overfitting, we employ a data augmentation procedure during training to each \ac{rd} map in the sequence, with identical transformations applied across all frames in the same sequence. Specifically, the following operations are 
applied:

\begin{itemize}
\item Addition of Gaussian noise with $\mu=0$ and \mbox{$\sigma=0.01$};
\item Flipping along the range or Doppler axis with probability $0.5$;
\item Random cyclic shift of $(s_r, s_d)$, where $s_r$ and $s_d$ are uniformly sampled from $(-s_{\text{max}}, s_{\text{max}})$ with $s_{\text{max}}=16$;
\item Random scaling by a factor uniformly sampled from $(0.9, 1.1)$.
\end{itemize}
After augmentation, the processed frames are reassembled into sequences and provided as input to the network. An overview of the complete pre-processing pipeline is shown in Figure~\ref{fig:preproc}.

\subsection{Proposed Architectures for Gesture Recognition}

We explore three distinct neural architectures for the radar-based gesture recognition task, each designed to process spatio-temporal information in two sequential stages:
\begin{enumerate}
    \item \textit{Spatial Feature Extraction:} In all three architectures, spatial features are extracted from the input frames using a pre-trained \ac{cnn} backbone, similarly to previous works on this dataset. 
    However, due to the relatively small amount of output classes, we opted for a more compact model, namely SqueezeNet 1.0~\cite{iandola2016squeezenet}, pre-trained on the ImageNet-1K dataset. This choice provides a favorable trade-off between model size and representational capacity. Preliminary experiments with custom \ac{cnn}s trained from scratch yielded inferior performance and slower convergence, motivating our choice. The backbone is fine-tuned on our task to adapt to the specific characteristics of radar \ac{rd} maps. This module returns a sequence of latent vectors of dimension $512$, which becomes the input for the next network block.
    \item \textit{Temporal Feature Extraction}: The temporal processing stage distinguishes the proposed architectures, as described below.
\end{enumerate}

\paragraph{CNN Architecture}
In this configuration, temporal dependencies are modeled using two one-dimensional convolutional layers applied to the sequence of spatial features extracted by the \ac{cnn} backbone. The spatial feature dimension is treated as the channel dimension, while the two convolutional layers employ kernel sizes of $[5,3]$, with $[256,128]$ output channels respectively, each followed by a ReLU activation. This design allows the network to capture both short- and medium-scale temporal patterns. 
The resulting features are then average-pooled along the time dimension and fed into a compact \ac{mlp}, consisting of a hidden layer with $128$ units and a final output layer for classification. 
The model is trained using the cross-entropy loss function. 
We will refer to this architecture as CNN2D+1D.
\paragraph{LSTM Architecture}
Here, the temporal modeling is performed by a small \ac{rnn}, equipped with a single layer of \ac{lstm} units, which are well-suited for capturing long-term dependencies in sequential data. The hidden feature size is set to $128$, balancing expressiveness and computational efficiency. The \ac{lstm} output is passed to a final dense output layer for classification. We will refer to this model as CNN+LSTM.

\paragraph{GRU Architecture}
This variant follows the same structure of the \ac{lstm} Architecture, but employs \ac{gru} cells instead. 
This architecture will be denoted as CNN+GRU.

\paragraph{SNN Architecture}
This architecture performs temporal modeling of spatial features with a lightweight \ac{snn}, to investigate energy-efficient temporal processing based on biology-inspired computation. We employ two hidden layers, with $128$ and $64$ neurons, respectively, composed of \ac{lif} neurons, which model temporal dynamics through membrane potential decay and discrete spike events. Two hidden layers are used, with $128$ and $64$ neurons, respectively, followed by a dense output layer. The spiking module takes the sequence of spatial feature vectors, coming from the SqueezeNet backbone, as a direct current input, so no spike-encoding mechanism is employed. This is an example of an hybrid architecture, mixing standard floating point computations with sparse, spike-based computations, and we will refer to it as CNN+SNN, or simply as the hybrid architecture.

\subsection{Training details}
All the proposed architectures are trained end-to-end, jointly optimizing the spatial backbone and the temporal module. 
For the hybrid architecture, surrogate gradients are employed using an $\text{arctan}$ function to approximate the gradients~\cite{eshraghian2023training}. 
All models are trained for $30$ epochs using the ADAM optimizer, and a mini-batch size of $16$. The three \ac{ann} models use a standard cross-entropy loss function, while the hybrid model employs a Cross Entropy Spike Rate Loss \cite{eshraghian2023training}. This criterion treats the spike count of output neuron $i$, defined as $c_i$, as a logit value, i.e., the output probability for class $i$ is obtained as:

\begin{equation}
    p_i = \frac{e^{c_i}}{\sum_{i=1}^{M}e^{c_i}},
\end{equation}
and the final loss value is computed with the usual cross-entropy function:

\begin{equation}
\label{eq:cesr}
    \mathcal{L}_{\rm CESR} = \sum_{i=1}^{M}y_i \log{p_i},
\end{equation}
where $\mathbf{y} = \{y_i\}_{i=1}^{M}$ is the one-hot encoded target vector. 

Finally, in the hybrid model, both the threshold and decay parameters of the \ac{lif} neurons are learned via surrogate gradient descent. All the models are trained on the native training split of the Aircraft Marshalling dataset, which is in turn divided into training and validation sets with a $90\%-10 \%$ ratio.
All the models are trained on an NVIDIA RTX 3080 GPU, using the PyTorch \cite{paszke2019pytorch} and SNNTorch \cite{eshraghian2023training} Python frameworks.

\section{Experimental Results}


\subsection{Performance Evaluation and Discussion}
\label{sec:experiments_discussion}

In this section, we present and discuss the achieved experimental results.
Our results are compared against two existing approaches addressing the same task on the same dataset~\cite{10149465,10734929}. The evaluation considers classification accuracy, memory footprint, and computational complexity to provide a comprehensive performance assessment and highlight the associated trade-offs. 

\begin{figure*}[t]
    \centering
    \begin{subfigure}[b]{0.366\textwidth}
        \centering
        \includegraphics[width=\textwidth]{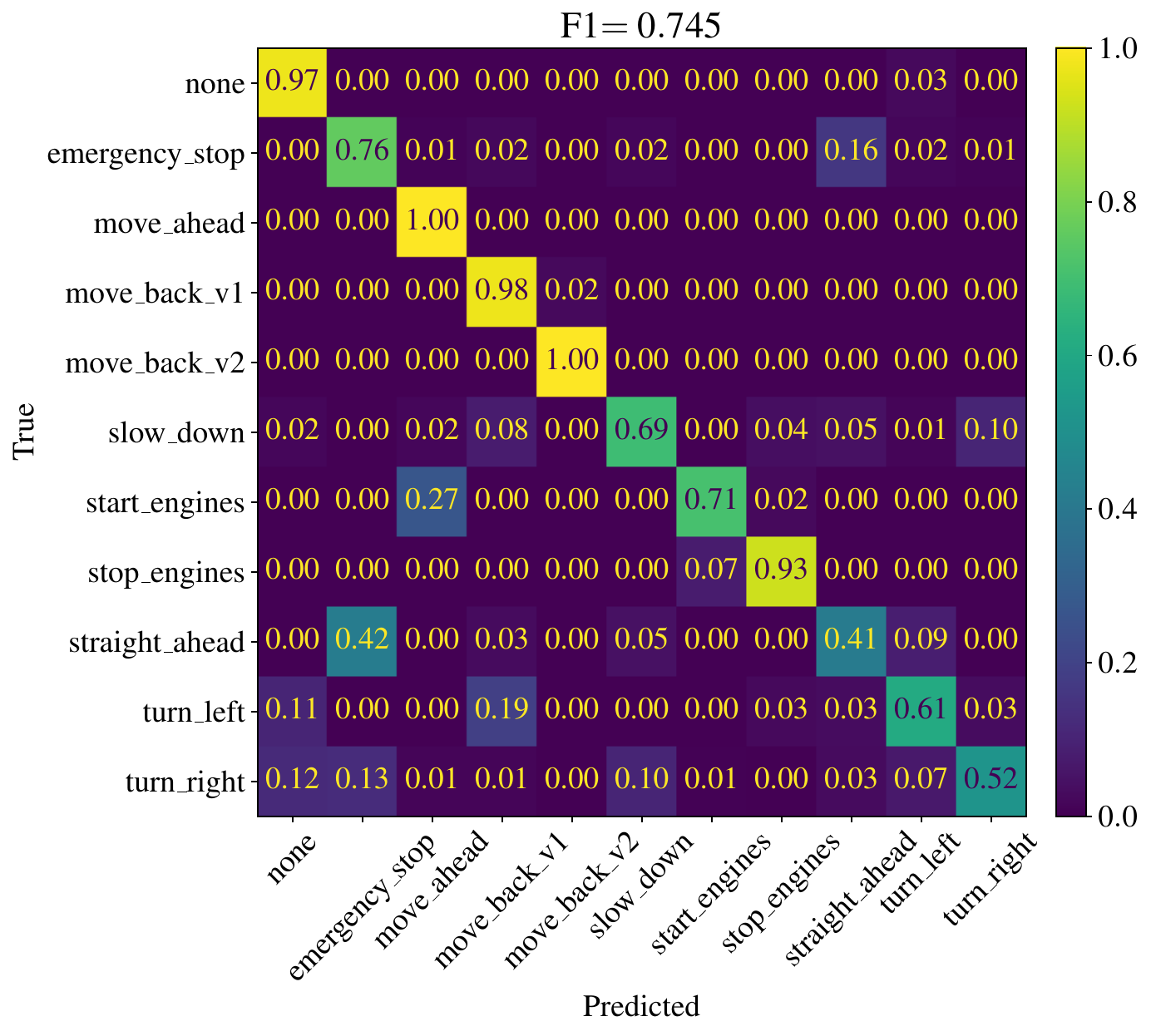}
        \caption{\ac{cnn} model}
        \label{fig:cfmtrx_cnn}
    \end{subfigure}%
    \hfill
    \begin{subfigure}[b]{0.312\textwidth}
        \centering
        \includegraphics[width=\textwidth]{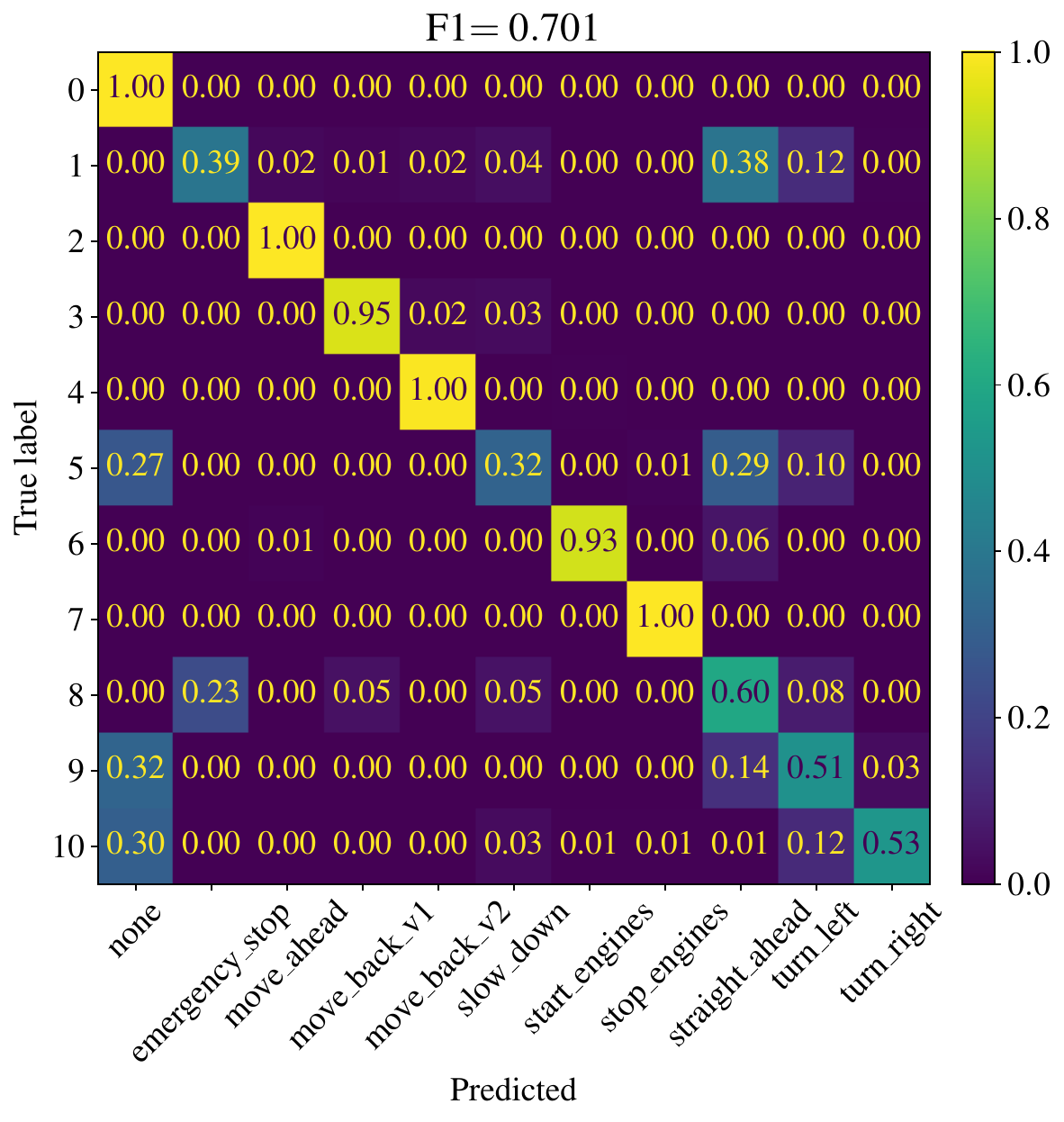}
        \caption{\ac{lstm} model}
        \label{fig:cfmtrx_lstm}
    \end{subfigure}%
    \hfill
    \begin{subfigure}[b]{0.312\textwidth}
        \centering
        \includegraphics[width=\textwidth]{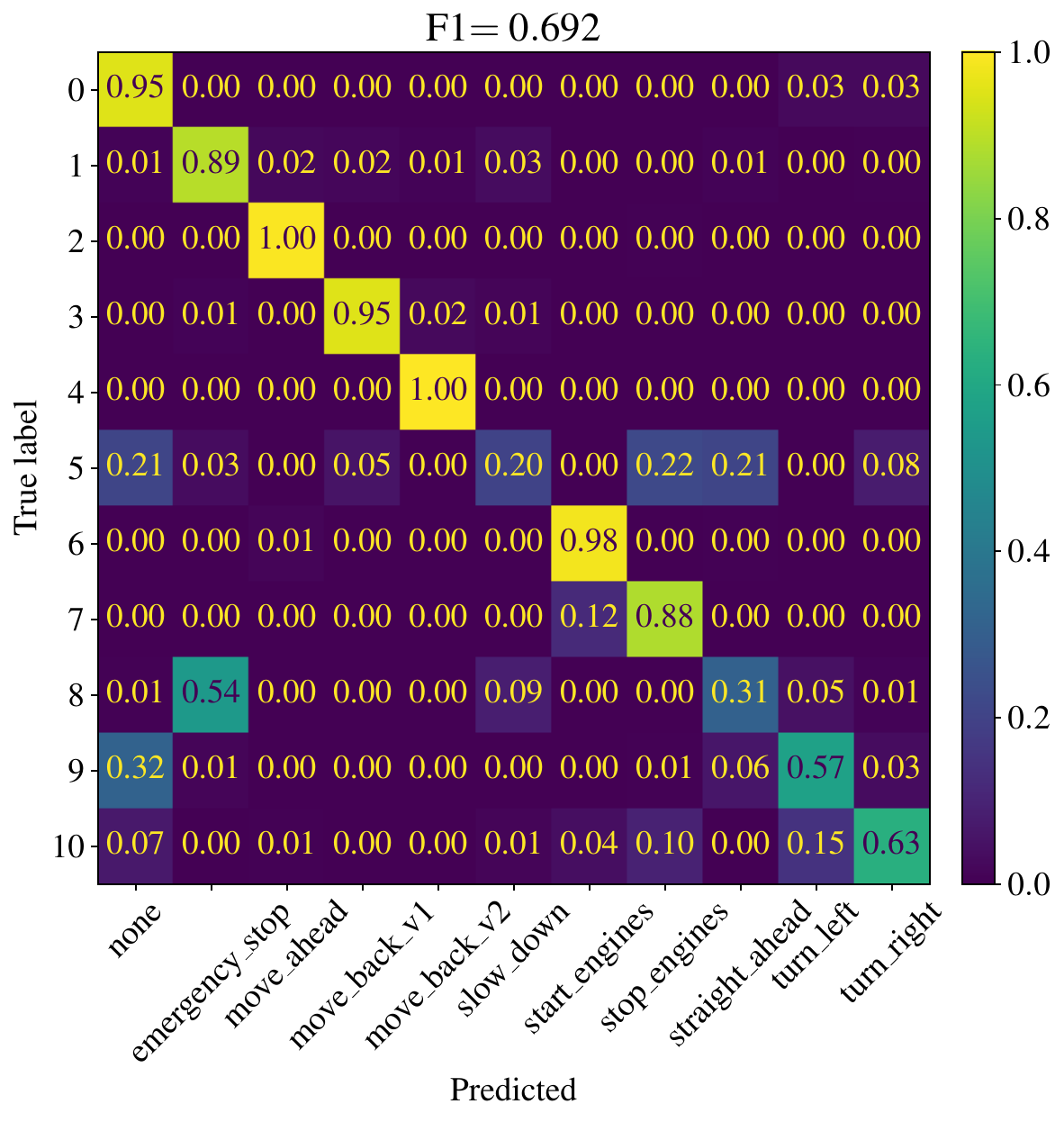}
        \caption{Hybrid model}
        \label{fig:cfmtrx_hyb}
    \end{subfigure}%

    \caption{Example of confusion matrices showing runs with average performance across the 10 performed runs, for each model type.}
    \label{fig:conf_matrices}
\end{figure*}

In Table \ref{tab:results} we report our results in terms of 10-fold cross validation accuracy and F$1$ score on the test set. Since this is a multi-class classification problem, the F$1$ score is computed for each class and then aggregated using the \textit{macro} averaging strategy \cite{manning2008introduction}, which assigns equal weight to all classes irrespective of their sample sizes.
More in detail, given $M$ the total number of classes, $\text{TP}_i$, $\text{FP}_i$, and $\text{FN}_i$ the true positives, false positives, and false negatives for class $i$, respectively, the macro-averaged F1 score is computed as follows:
\begin{equation}
\text{F$1$} = \frac{1}{M} \sum_{i=1}^{M} \frac{2\text{TP}_i}{2\text{TP}_i + \text{FP}_i + \text{FN}_i}.
\end{equation}
For the baselines, we report the performances from the original papers \cite{10149465, 10734929}. 
For the memory cost of each model, we report the size in MB of both input tensors and learnable parameters, while for the computational complexity we report the number of \ac{flops} required to perform a forward pass for a single input frame, to provide a direct comparison with the existing single-frame approaches.
In Figure~\ref{fig:conf_matrices}, we also report some example confusion matrices, showcasing the performance of the proposed models on the test set.

Our results demonstrate that the CNN2D+1D architecture achieves the highest accuracy, surpassing the current state of the art on the Aircraft Marshalling Signals benchmark. This provides further evidence that the employed sequential input representation is the best choice for \ac{har} tasks. Moreover, given the significantly reduced size of the \ac{cnn} backbone, we are able to reduce the number of trainable parameters, thus decreasing the overall memory footprint of the model. This observation holds consistently across all the explored architectures, where the memory footprint is reduced by a factor ranging from 5 to 8, compared to the baselines.

The proposed architectures also require fewer FLOPs to process a single input frame, approximately half those of the baseline models. However, we highlight that our input consists of 15 frames, while the baselines use only one. Therefore, despite the more compact architecture, the employed sequential input representation yields an overall higher cost for a full forward pass in terms of theoretical FLOPs, at least without any further optimization (which will be the subject of Section \ref{sec:pruning}). 

While the CNN2D+1D network achieves the best performance, it also has the largest input memory footprint, as the entire input sequence must be stored in memory before producing a prediction. However, due to their recurrent structure, the other architectures can afford to process one frame at a time, thus saving on some allocated input memory.

The hybrid CNN+SNN model does not achieve the top accuracy, though its performance remains comparable to both the other proposed models and the baselines. However, it offers several advantages over more conventional architectures. First, it exhibits the lowest memory footprint among all the proposed models. This is due to the spiking nature of the \ac{snn} architecture, which employs \ac{lif} neurons to leverage internal membrane potential states, together with learnable decay factors and thresholds, to capture temporal correlations in the input. These mechanisms result in a more compact memory footprint compared to traditional RNN-based approaches.

Additionally, the loss function used to train the hybrid model, defined in Eq.~(\ref{eq:cesr}), provides it with an additional capability compared to the other architectures. 
Due to their architecture design, \ac{cnn} and \ac{rnn} based models require the full input sequence to produce a prediction.
On the other hand, a \ac{snn} trained with a cross-entropy spike rate criterion learns to elicit a higher spike count at the output neuron corresponding to the predicted class.

\begin{figure}
    \centering
    \includegraphics[width=0.75\linewidth]{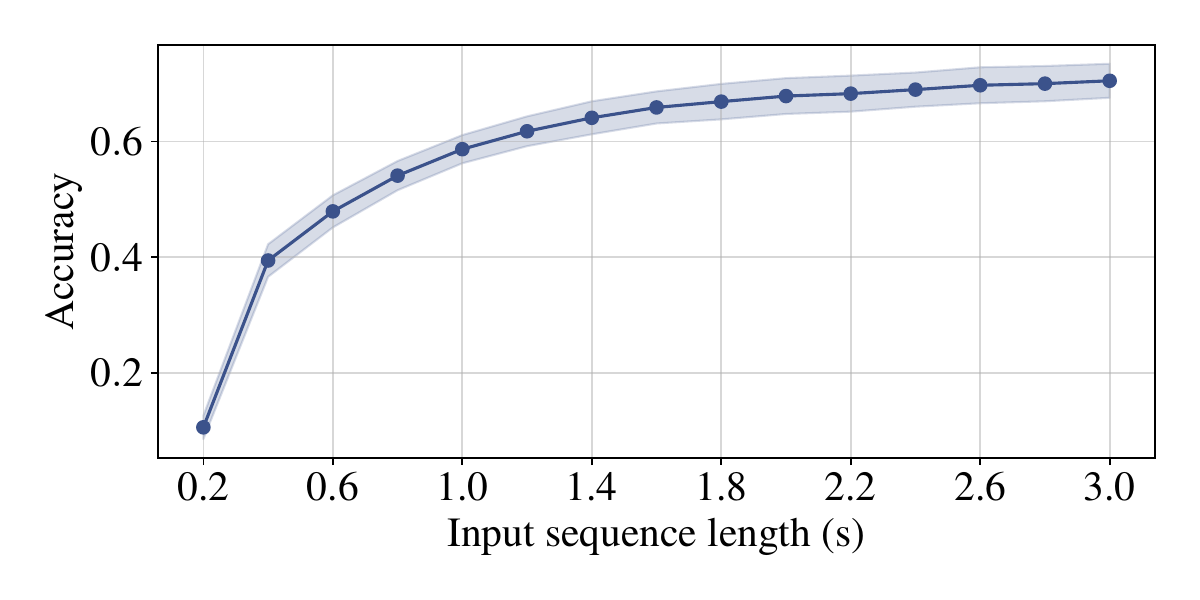}
    \caption{Latency accuracy trade-off in the hybrid model.}
    \label{fig:latency}
\end{figure}

Notably, this property enables a controllable trade-off between prediction latency and accuracy: at inference time, the system can generate a prediction from only a fraction of the input sequence, allowing faster responses at the cost of reduced confidence. Additionally, by processing fewer input frames, we can mitigate the overall computational complexity for a full forward pass. 
This kind of flexibility is highly desirable for edge applications, where constraints on latency and energy consumption are critical. 
Figure~\ref{fig:latency} illustrates this trade-off, showing a non-linear, concave relationship between latency and accuracy. Accuracy rises rapidly within the first few timesteps and then increases more gradually beyond steps~6--7. This is a desirable behavior for the aforementioned edge-oriented use cases.

\begin{table*}[]
\caption{Results of all models in terms of Accuracy and F1-Score (macro-average) on the Aircraft Marshalling Gesture dataset. Results are reported as the average and standard deviation of 10 different random seeds.}
\centering
\resizebox{\linewidth}{!}{\begin{tabular}{ccccccc}
\toprule
\textbf{Model}      & \textbf{Acc. (\%)} & \textbf{F1-Score} & \textbf{Params (MB)} & \textbf{Input (MB)} & \textbf{FLOPs (1 frame)} \\ \midrule 
Baseline 1 \cite{10149465} & $64.6$    & -  & $24.90$           & $0.063$ & $6.90\cdot 10^8$      \\
Baseline 2 \cite{10734929} & $71.4$    & - & $25.86$           & $0.063$ & $6.90\cdot 10^8$     \\ \midrule
CNN2D+1D  & $\mathbf{72.99\pm 3.39}$    & $\mathbf{0.694 \pm 0.033}$ & $5.90$           & $0.945$ & $3.68 \cdot 10^8$     \\
CNN+LSTM  & $69.91 \pm 1.71$    & $0.689 \pm 0.015$   & $4.03$           & $0.063$  & $3.68\cdot10^8$    \\
CNN+GRU   & $68.20\pm 3.44$    & $0.663\pm 0.042$ & $3.72$  & $0.063$ & $3.68\cdot10^8$  \\
CNN+SNN & $70.57\pm2.82$    & $0.686\pm0.032$    & $\mathbf{3.05}$           & $0.063$ & $3.68\cdot10^8$ \\ 
\bottomrule   
\end{tabular}}


\label{tab:results}
\end{table*}

The proposed architectures are designed and validated for gestures of aircraft marshalling signals, but they can also be adapted to any kind of \ac{har} application employing sequential input radar representations. To show the generality and flexibility of the discussed models, we also replicated our experiments on a different popular dataset of hand gesture recognition, the Soli dataset \cite{wang2016interacting}. Performance metrics are reported in Table \ref{tab:resultssoli}. In this case, we used input sequences of approximately two seconds, as is common practice in the literature \cite{tsang2021radar, wang2016interacting}. Given the higher temporal resolution of the Soli radar sensor (40 Hz), this corresponds to sequences of 80 frames. All the models achieve high levels of accuracy proving their flexibility for \ac{har} tasks. 

\begin{table}[]


\caption{Results of Accuracy and F1-Score (macro-average) on the Soli Dataset.}
\centering
\resizebox{0.45\linewidth}{!}{\begin{tabular}{ccccccc}
\toprule
\textbf{Model}     & \textbf{Acc. (\%)} & \textbf{F1-Score} \\ \midrule 
CNN+2D1D  & $97.27 \pm 0.94$   & $0.974 \pm 0.01$  \\
CNN+LSTM  & $96.24 \pm 1.08$    & $ 0.964\pm 0.01$  \\
CNN+GRU  & $96.52 \pm 0.96$    & $ 0.965\pm 0.01$  \\
CNN+SNN  & $94.80 \pm 1.61$   & $0.949 \pm 0.002$ \\ 
\bottomrule   
\end{tabular}}

\label{tab:resultssoli}
\end{table}

\subsection{Model Pruning}
\label{sec:pruning}

As observed in Section \ref{sec:experiments_discussion}, despite a significant reduction in memory footprint and the added possibility for accuracy-latency trade-off, our hybrid \ac{snn} still requires a higher number of FLOPs for a full forward pass, compared to existing single-frame based approaches. However, additional optimization strategies can be applied to reduce the resulting computational costs.

\begin{figure}
    \centering
    \includegraphics[width=0.7\linewidth]{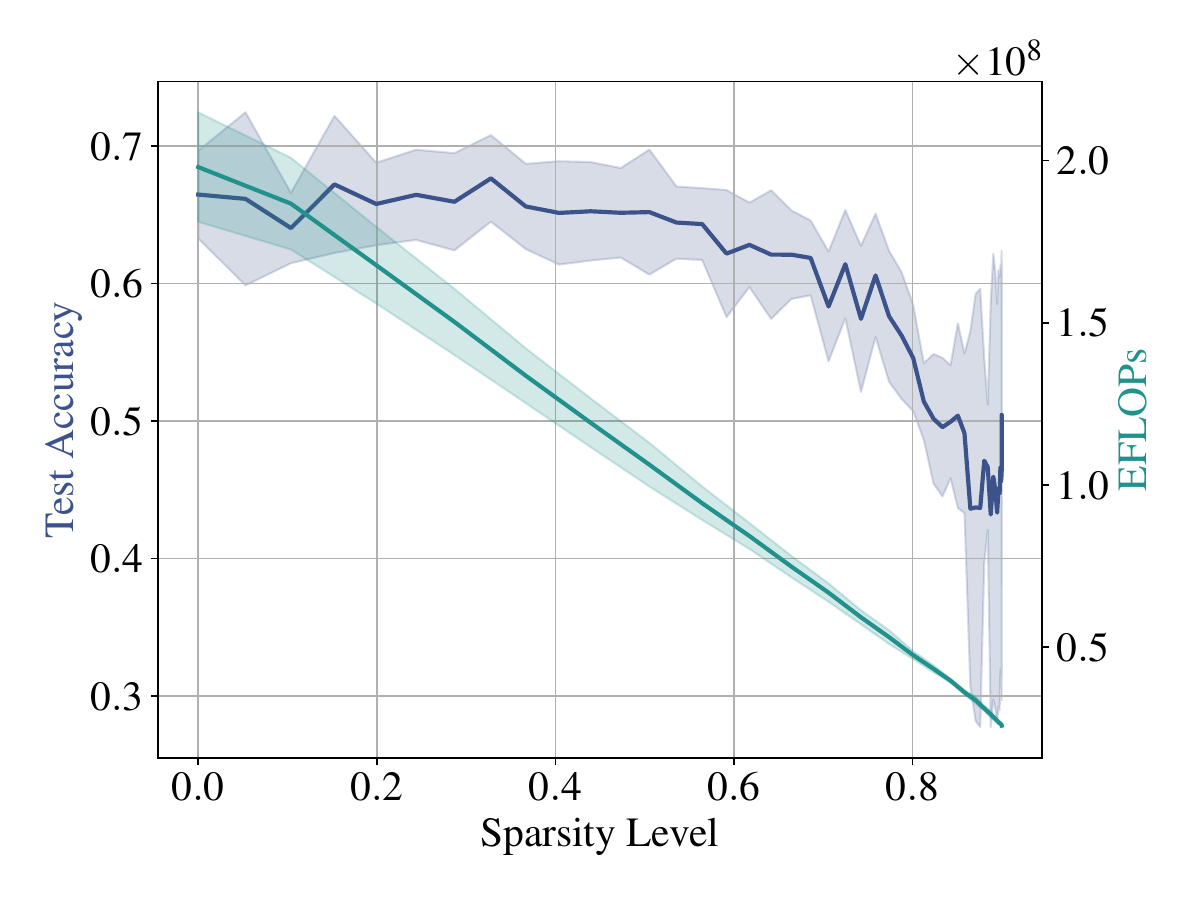}
    \caption{Depiction of the pruning process of our hybrid model. The figure shows how the accuracy on the test set is impacted as the sparsity of the network parameters increases. The shaded area shows the standard deviation dispersion measure.}
    \label{fig:pruning}
\end{figure}

Sparse computation is a key factor in improving the efficiency of neural network models, as it reduces both memory usage and computational cost. A common approach to induce sparsity is pruning, a practice that consists of removing parameters with minimal contribution to the model’s output. In this work, we adopt the pruning strategy proposed in~\cite{zhu2017prune} to further optimize the memory footprint of the hybrid model. Starting from a pre-trained model, a secondary training stage is performed in which, after a fixed number of iterations, the weights with the smallest magnitudes are pruned. The network is then fine-tuned to recover performance before the next pruning step. This iterative process provides fine-grained control over the target sparsity level while enabling the remaining weights to adapt and compensate for the removed connections. The outcomes of these pruning experiments are presented in Figure~\ref{fig:pruning}.

Beyond computational complexity, model sparsity also impacts energy efficiency, a crucial aspect for resource-constrained deployments. Evaluating energy efficiency is inherently challenging, as it strongly depends on the target hardware platform, and FLOPs are not a reliable metric to provide an estimate. To provide a hardware-agnostic yet informative assessment, we estimate the energy efficiency of the pruned hybrid architecture by computing 
\acp{eflop} a recently proposed metric that accounts for sparsity in the computation of FLOPs to provide a more realistic measure of energy consumption~\cite{narduzzi2025eflop}. More in detail, for each sparsity level reached during the pruning procedure, we compute \acp{eflop} for a forward pass of a single frame through the network, proceeding as follows. For each FLOP:
\begin{enumerate}
    \item if it is a multiplication, we exclude it from the count if either of the two operands is 0;
    \item if it is an addition, we exclude it from the count if both operands are 0s.
\end{enumerate}

We remark that \ac{eflop} is a data-dependent metric of computational complexity, since also the sparsity of the input will influence the outcome. For this reason, we report confidence intervals also for this metric.
In Figure~\ref{fig:pruning}, we show the amount of \acp{eflop} as the weight sparsity of the model varies during the pruning procedure, showing a linear dependency. 
Remarkably, the hybrid architecture is capable of retaining near-optimal levels of performance even at high sparsity levels, exhibiting a high potential for hardware optimizations once deployed on a dedicated edge-device.

\section{Discussion \& Conclusion}

In this work, we investigated the use of \acp{snn} for radar-based \ac{har}, with a particular focus on the Aircraft Marshalling Signals dataset. Leveraging a time-stepped radar input representation, we designed and validated network architectures capable of modeling temporal correlations across frames, surpassing the state of the art on this challenging benchmark. We further validated the proposed models on the Soli dataset, demonstrating their flexibility for general radar-based \ac{har} applications.

For the first time on this dataset, we introduced a hybrid \ac{snn} that combines a pre-trained convolutional backbone for spatial feature extraction with densely connected \ac{lif} neurons for temporal modeling, achieving performance comparable to state-of-the-art \ac{ann}-based approaches. Our analysis highlights the unique trade-offs enabled by spiking computation. Our hybrid model achieves an 88\% reduction in trainable parameters compared to baseline approaches, with an accuracy drop of less than one percentage point, while also supporting a tunable balance between prediction latency and accuracy. These properties make \ac{snn}-based solutions particularly attractive for resource-constrained edge scenarios where efficiency and adaptability are critical.

Future research directions include the development of fully spiking architectures for radar-based \ac{har}, to fully exploit the advantages of event-driven computation. Preliminary experiments showed that training such models from scratch was infeasible, and we observed that a pre-trained convolutional backbone was necessary to achieve generalization on the domain-shifted test set. This limitation can likely be attributed to two main factors: the sub-optimal training strategies currently available for \acp{snn}, which rely on surrogate gradients, and the relatively small size of the dataset, despite the employed augmentation strategies. Alternative approaches, such as \ac{ann}-to-\ac{snn} conversion, may provide a more practical path toward fully spiking implementations. A fully spiking architecture could also enable deep sensor fusion of radar and \ac{dvs} camera data, the latter providing a natively spike-based representation, available in the Aircraft Marshalling dataset. An end-to-end spiking framework capable of jointly processing both modalities could deliver further improvements in recognition accuracy while retaining the benefits of spiking computation.

\vspace{1cm}

\bibliographystyle{IEEEtran}
\bibliography{IEEEabrv,biblio}

\end{document}